\documentclass[letterpaper, 10 pt, conference]{ieeeconf}  % Comment this line out if you need a4paper

\IEEEoverridecommandlockouts                              % This command is only needed if 
\usepackage{amsmath} % assumes amsmath package installed
\usepackage{subcaption}
\usepackage{float}
\usepackage{todonotes}
\usepackage{amsmath}
\usepackage{amssymb}
\usepackage{diagbox}
\usepackage{makecell}
\usepackage{amsmath}
\usepackage{url}
\usepackage{hyperref}
\usepackage{bm}
\usepackage{array}
\let\labelindent\relax
\usepackage{enumitem}
\usepackage{color,soul}
\usepackage{leftindex}
\usepackage{graphicx}
\usepackage{tensor}   % For tensor notation with better spacing
\usepackage{stackengine}

\usepackage{adjustbox}
\usepackage{threeparttable}
\usepackage{multirow}
\usepackage{booktabs}

\title{\LARGE \bf
Aerial Manipulation in the Wild with Onboard Perception, Policy Learning, and Whole-Body Control
}

\author{Yuanzhu Zhan$^{*1}$, Yufei Jiang$^{*1}$, Zemu Zhang$^{1}$, Junyi Geng$^{1}$
\thanks{$^{*}$ Equal contribution.}
\thanks{$^{1}$ Department of Aerospace Engineering, Pennsylvania State University, University Park, PA, 16802, USA. 
{\tt\footnotesize \{yvz6008, yufei\_jiang, zqz5459, jgeng\}@psu.edu}}%
}

\begin{document}

\maketitle
\thispagestyle{empty}
\pagestyle{empty}

%%%%%%%%%%%%%%%%%%%%%%%%%%%%%%%%%%%%%%%%%%%%%%%%%%%%%%%%%%%%%%%%%%%%%%%%%%%%%%%%
\begin{abstract}

Aerial manipulation in outdoor environments remains challenging due to the simultaneous requirements of reliable state estimation, stable aerial motion, and precise manipulation under external disturbances. In this work, we present a real-world outdoor aerial manipulation framework that integrates imitation learning, onboard LiDAR-inertial state estimation, and whole-body model predictive control. A Diffusion Policy is trained from manipulation demonstrations to generate desired end-effector motions from onboard observations. These learned commands are executed by a whole-body MPC that jointly coordinates the aerial platform and manipulator to realize the desired end-effector trajectory. To eliminate reliance on external motion-capture infrastructure, the platform employs onboard LiDAR-inertial odometry for state estimation during outdoor operation. We validate the complete framework on a physical aerial manipulator and demonstrate successful execution of outdoor manipulation tasks. The experimental results show that demonstration-driven manipulation policies can be effectively integrated with onboard state estimation and model-based whole-body control to enable aerial manipulation beyond controlled indoor environments.

\end{abstract}

%%%%%%%%%%%%%%%%%%%%%%%%%%%%%%%%%%%%%%%%%%%%%%%%%%%%%%%%%%%%%%%%%%%%%%%%%%%%%%%%
\section{Introduction}
\label{sec: intro}

Traditional unmanned aerial vehicles (UAVs) have demonstrated strong capabilities in autonomous navigation, sensing, inspection, and data collection in environments that are difficult or unsafe for humans to access~\cite{jiang2025selfsupervised}. Aerial manipulation extends these capabilities from passive observation to physical interaction by equipping UAVs with robotic manipulators or specialized end effectors~\cite{ruggiero2018aerial,ollero2021past,zhan2026contactaware}. However, manipulation introduces additional challenges because arm motion affects the floating aerial base, while the vehicle must simultaneously maintain stable flight and accurately control the end effector~\cite{yang2014dynamics,kim2018cooperative}. These challenges are further amplified outdoors, where high-precision state estimation cannot rely on laboratory motion-capture infrastructure and environmental conditions are less controlled.

Imitation learning provides an alternative to manually designing task-specific manipulation behaviors. Visuomotor policies can learn directly from demonstrations~\cite{levine2016endtoend,zhao2023act}, while Diffusion Policy has shown strong performance in generating temporally coherent manipulation motions~\cite{chi2023diffusion}. Recent works such as UMI-on-Legs, UMI-on-Air, and Flying Hand have demonstrated the potential of demonstration-driven policies for mobile and aerial manipulation~\cite{ha2024umilegs,gupta2025umionair,he2025flying}. However, these learning-based aerial manipulation systems have largely been evaluated in controlled indoor environments.

Compared with controlled indoor settings, outdoor deployment introduces additional challenges including less structured visual conditions, stronger environmental disturbances, and the absence of laboratory-grade localization infrastructure. Existing systems typically address these challenges through carefully designed task-specific perception and control modules~\cite{ramonsoria2020outdoor,ubellacker2024highspeed,bauer2025wild}, but such pipelines do not directly address whether manipulation behaviors learned from demonstrations can transfer across environments.

Motivated by this gap, we investigate whether a visuomotor policy trained solely from indoor demonstrations can be deployed for outdoor aerial manipulation without retraining. We develop and experimentally evaluate a framework that integrates Diffusion Policy, LiDAR-inertial state estimation, and whole-body MPC for real-world outdoor aerial manipulation. We validate the complete system on a physical aerial manipulator and demonstrate successful outdoor manipulation using a policy trained entirely from indoor demonstrations.

The main contributions of this work are:
\begin{itemize}
    \item An integrated framework combining diffusion-policy-based imitation learning, onboard LiDAR-inertial state estimation, and whole-body MPC for outdoor aerial manipulation.
    
    \item A whole-body MPC formulation for executing learned end-effector motions through coordinated aerial-base and manipulator motion.
    
    \item Real-world outdoor validation without external motion capture, outdoor demonstrations, or policy retraining.
\end{itemize}
% \section{Related Works}
\label{sec: related works}
\section{Methodology}
\label{sec: methodology}

\subsection{System Overview}
\label{subsec: sys_overview}
% Our system comprises a handheld UMI device for human demonstration collection and an aerial manipulator for policy deployment, as shown in Fig.~\ref{fig:sys_overview}. Both platforms use an OAK-1W egocentric camera and identical soft-finger grippers, providing a consistent visuomotor interface between demonstration and execution. The handheld device additionally uses motion-capture markers for pose tracking and fiducial markers for gripper width estimation during indoor data collection. (The built-in visual-inertial odometry in the camera can also support in-the-wild data collection.)

\begin{figure}[t]
    \centering

    \begin{subfigure}{0.75\linewidth}
        \centering
        \includegraphics[width=\linewidth]{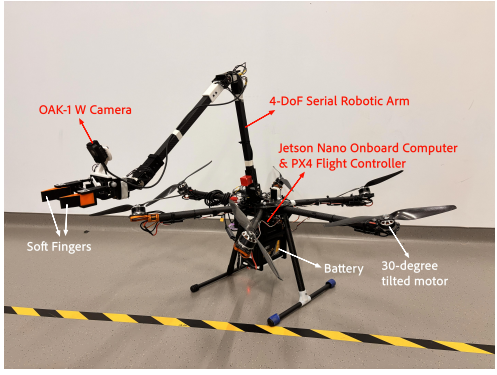}
        \caption{AM platform}
        \label{fig:AM_platform}
    \end{subfigure}
    \hfill
    
    \begin{subfigure}{0.75\linewidth}
        \centering
        \includegraphics[width=\linewidth]{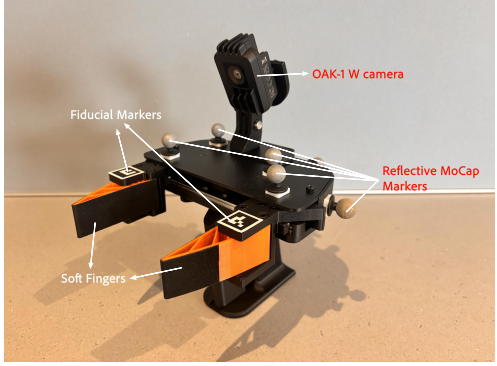}
        \caption{UMI handheld gripper}
        \label{fig:UMI_mini}
    \end{subfigure}
    \caption{System Overview}
    \label{fig:sys_overview}
\end{figure}

Our aerial manipulation system consists of a fully actuated hexarotor, a 4-DoF serial manipulator, and an end-effector equipped with an OAK-1W camera, as shown in Fig.~\ref{fig:sys_overview}. During deployment, the visuomotor policy generates end-effector commands, while onboard LiDAR–inertial odometry provides floating-base state feedback and a whole-body controller coordinates the aerial base and manipulator.

% The aerial platform consists of a fully-actuated hexarotor, a 4-DoF serial arm, and the corresponding camera-gripper setup. During deployment, the learned policy generates task-space end-effector commands, while onboard LIO provides the floating drone base state required by the whole-body controller. This enables the same indoor-trained policy to be deployed outdoors without external motion capture system or policy retraining.

For policy training, demonstrations are collected offline using a handheld UMI device~\cite{gupta2025umionair}. This shared end-effector-centric interface facilitates transferring the learned policy from handheld demonstrations to the aerial manipulator.

% \begin{figure}[t]
%     \centering
%     \includegraphics[width=0.9\linewidth]{Figures/AM_platform.pdf}
%     \caption{System Overview}
%     \label{fig:sys_overview}
% \end{figure}

\subsection{Diffusion Policy for End-Effector Action Generation}

Diffusion Policy formulates visuomotor control as conditional generative modeling of future action trajectories. During training, noise is progressively added to demonstrated action sequences, and a conditional denoising network learns to recover the original actions given the observation history. During inference, the policy starts from a noisy action sequence and iteratively denoises it to generate a future trajectory that is consistent with the current observations. Following DDIM~\cite{song2021DDIM}, the denoising process can be written as
\begin{equation}
    \mathbf{a}^{k-1} = \mathbf{a}^k + \psi_k(\pi_\theta(\mathbf{a}^k,t|\mathcal{O})),
\end{equation}
where $\mathbf{a}^k$ denotes the noisy action trajectory at diffusion step $k$, $\pi_\theta$ is the trained denoiser, $\mathcal{O}$ is the conditioning observation, and $\psi_k$ is the DDIM update function for step $k$.

For aerial manipulation, we adopt the end-effector-centric representation used in~\cite{chi2024umi,gupta2025umionair}. At each inference step, the policy receives a temporal observation window containing egocentric RGB images $\mathbf{I}$, relative end-effector poses $\mathbf{T}^\text{ee}$, and gripper widths $w$, so the observation $\mathcal{O}$ is defined as:
\begin{equation}
    \mathcal{O} = \{ \mathbf{I}_t, \mathbf{T}^\text{ee}_t, w_t\}_{t=1}^L,
    \label{eq:obs}
\end{equation}
where $L$ denotes the observation horizon. The policy predicts a horizon of future end-effector actions represented by relative end-effector positions $\mathbf{p}^\text{ee}$, orientations $\mathbf{R}^\text{ee}$, and gripper-width commands $w$,
\begin{equation}
    \mathbf{a} = \{\mathbf{p}_t^\text{ee}, \mathbf{R}_t^\text{ee}, w_t\}_{t=1}^N,
    \label{eq:action}
\end{equation}
where $N$ denotes the planning horizon.

Because the policy operates in end-effector space rather than directly commanding the aerial base, manipulator joints, or actuators, the end-effector-centric architecture decouples the learned policy from the platform-specific kinematics and dynamics, which are handled by the downstream whole-body controller. The relative pose representation further reduces the policy's dependence on precise global localization.

Human demonstrations are collected using the handheld UMI device described in Sec.~\ref{subsec: sys_overview}, whose camera–gripper configuration closely matches that of the aerial end-effector. The collected demonstrations are converted into the observation–action sequences defined in (\ref{eq:obs}-\ref{eq:action}) and used to train the conditional UNet-based~\cite{ronneberger2015unet} Diffusion Policy.

\subsection{Whole-Body Model Predictive Control}

To execute the end-effector trajectories generated by the learned policy, we employ a whole-body model predictive controller that jointly coordinates the aerial platform and manipulator. The state and control vectors are defined as
\begin{equation}
\begin{aligned}
\mathbf{x} &=
\begin{bmatrix}
\mathbf{p}^{\top} &
\mathbf{v}^{\top} &
\boldsymbol{\eta}^{\top} &
\boldsymbol{\omega}^{\top} &
\mathbf{q}^{\top}
\end{bmatrix}^{\top}, \\
\mathbf{u} &=
\begin{bmatrix}
\mathbf{f}^{\top} &
\boldsymbol{\tau}^{\top} &
\mathbf{q}_{\mathrm{ref}}^{\top}
\end{bmatrix}^{\top},
\end{aligned}
\label{eq:mpc_state_control}
\end{equation}
where $\mathbf{p},\mathbf{v}\in\mathbb{R}^{3}$ are the aerial-base position and linear velocity, $\boldsymbol{\eta},\boldsymbol{\omega}\in\mathbb{R}^{3}$ are the base Euler angles and angular velocity, and $\mathbf{q}\in\mathbb{R}^{4}$ denotes the manipulator joint configuration. The control inputs $\mathbf{f},\boldsymbol{\tau}\in\mathbb{R}^{3}$ are the commanded force and torque, and $\mathbf{q}_{\mathrm{ref}}\in\mathbb{R}^{4}$ is the desired joint configuration.

At each control cycle, the MPC solves the finite-horizon nonlinear optimal control problem
\begin{equation}
\begin{aligned}
\min_{\mathbf{x}_{0:N},\mathbf{u}_{0:N-1}}
\quad &
\sum_{k=0}^{N-1}
\ell_k(\mathbf{x}_k,\mathbf{u}_k)
+
\ell_N(\mathbf{x}_N) \\
\text{s.t.}\quad
&
\mathbf{x}_{k+1}=f(\mathbf{x}_k,\mathbf{u}_k), \\
&
\mathbf{x}_{\min}
\leq
\mathbf{x}_k
\leq
\mathbf{x}_{\max}, \\
&
\mathbf{u}_{\min}
\leq
\mathbf{u}_k
\leq
\mathbf{u}_{\max}, \\
&
\mathbf{x}_0=\mathbf{x}_{\mathrm{meas}} .
\end{aligned}
\label{eq:mpc_ocp}
\end{equation}

The state-transition function $f(\mathbf{x}_k,\mathbf{u}_k)$ is obtained by discretizing the following lightweight whole-body dynamics model:
\begin{equation}
\begin{aligned}
\dot{\mathbf{p}} &= \mathbf{v}, \\
\dot{\mathbf{v}} &= \frac{1}{m}\mathbf{f}, \\
\dot{\boldsymbol{\eta}} &\approx \boldsymbol{\omega}, \\
\dot{\boldsymbol{\omega}} &= \mathbf{I}^{-1}\boldsymbol{\tau}, \\
\dot{\mathbf{q}} &= k_q
\left(
\mathbf{q}_{\mathrm{ref}}-\mathbf{q}
\right),
\end{aligned}
\label{eq:mpc_dynamics}
\end{equation}
where $m$ is the total mass of the aerial manipulator, $\mathbf{I}=\mathrm{diag}(I_x,I_y,I_z)$ is the rotational inertia, and $k_q$ approximates the closed-loop response of the manipulator joint controller. A small-angle approximation, $\dot{\boldsymbol{\eta}}\approx\boldsymbol{\omega}$, is used for the base attitude dynamics.

The stage cost in~\eqref{eq:mpc_ocp} is defined as
\begin{equation}
\begin{aligned}
\ell_k =
&\,
\|\mathbf{p}_{E,k}-\mathbf{p}^{d}_{E,k}\|_{\mathbf{Q}_{p}}^2
+
\|\boldsymbol{\eta}_{E,k}-\boldsymbol{\eta}^{d}_{E,k}\|_{\mathbf{Q}_{\eta}}^2
\\
&+
\|\mathbf{v}_k\|_{\mathbf{Q}_{v}}^2
+
\|\boldsymbol{\eta}_k-\boldsymbol{\eta}^{d}_{B,k}\|_{\mathbf{Q}_{B}}^2
+
\|\boldsymbol{\omega}_k\|_{\mathbf{Q}_{\omega}}^2
\\
&+
\|\mathbf{q}_k-\mathbf{q}_{0}\|_{\mathbf{Q}_{q}}^2
+
\|\mathbf{u}_k-\mathbf{u}_{0}\|_{\mathbf{R}}^2
+
\|\mathbf{u}_k-\mathbf{u}_{k-1}\|_{\mathbf{R}_{\Delta}}^2 .
\end{aligned}
\label{eq:mpc_stage_cost}
\end{equation}
Here, $\mathbf{p}_{E,k}$ and $\boldsymbol{\eta}_{E,k}$ denote the end-effector position and orientation, while $\mathbf{p}^{d}_{E,k}$ and $\boldsymbol{\eta}^{d}_{E,k}$ are the corresponding references generated by the learned policy. The first two terms penalize end-effector tracking errors. The remaining state terms regularize base velocity, attitude, angular velocity, and manipulator configuration. The nominal arm configuration $\mathbf{q}_{0}$ encourages the manipulator to remain near a preferred posture, while the control and input-smoothness terms penalize excessive control effort and abrupt command changes. The desired base orientation $\boldsymbol{\eta}^{d}_{B,k}$ maintains a near-level vehicle attitude while allowing the yaw direction to follow the end-effector reference. The terminal cost $\ell_N$ consists of the end-effector tracking and state-regularization terms evaluated at the final prediction step, excluding the control-effort and input-smoothness penalties. 

The end-effector states appearing in~\eqref{eq:mpc_stage_cost} are obtained from the aerial-base state and manipulator configuration through the whole-body forward kinematics,
\begin{equation}
{}^{W}\mathbf{T}_{E,k}
=
\mathcal{F}\!\left(
\mathbf{p}_{k},
\boldsymbol{\eta}_{k},
\mathbf{q}_{k}
\right),
\label{eq:whole_body_fk}
\end{equation}
where ${}^{W}\mathbf{T}_{E,k}$ denotes the end-effector pose in the world frame, from which $\mathbf{p}_{E,k}$ and $\boldsymbol{\eta}_{E,k}$ are extracted.

The nonlinear optimal control problem is implemented in \texttt{acados} using a nonlinear least-squares objective and solved with sequential quadratic programming real-time iteration (SQP-RTI). At each control step, only the first optimized control input is applied in a receding-horizon manner. This formulation enables coordinated motion of the aerial base and manipulator for tracking the policy-generated end-effector trajectory.
\section{Experiments}
\label{sec: experiments}
We evaluate the proposed system to answer two questions: (1) whether onboard LIO provides sufficiently reliable state feedback for outdoor aerial manipulation, and (2) whether a policy trained exclusively from indoor demonstrations can transfer to an unseen outdoor environment without retraining.

\subsection{Experimental Setup}
We evaluate the system on an aerial peg-in-hole task in which the vehicle must insert a 20 mm diameter stick into a 50 mm diameter hole positioned approximately 130 cm above the ground. The visuomotor policy is trained entirely from 80 indoor demonstrations collected with the handheld UMI device. During demonstration collection, we apply domain randomization by varying the initial gripper position and the target hole location on the wall, exposing the policy to diverse approach and insertion configurations. The resulting policy is used unchanged for both indoor and outdoor evaluation. During outdoor deployment, the aerial platform relies on FAST-LIO2~\cite{xu2022fastlio2} for onboard LiDAR-inertial state estimation of the floating base without external motion-capture infrastructure. No outdoor demonstrations, policy fine-tuning, or environment-specific adaptation are used.

A trial is considered successful when the stick is inserted into the target hole without human intervention after policy execution begins. We conduct 5 trails indoors and 3 trails in an unseen outdoor grass field.

Several metrics are defined to quantify the controller's performance:
\begin{itemize}
    \item Normalized end-effector position tracking error (Norm. EE Error)
    
    % Let $\mathbf{p}_{\mathrm{ee}}(t)$ and $\mathbf{p}^{\mathrm{ref}}_{\mathrm{ee}}(t)$ denote the measured and commanded end-effector positions at timestep $t$. We define
    \begin{equation*}
        \rho_{\text{ee}}
        =
        \frac{1}{N_t}
        \sum_{t=1}^{N_t}
        \frac{
        \left\|
        \mathbf{p}^{ee}_t-\mathbf{p}^{ee,\text{des}}_{t}
        % \mathbf{p}^{\mathrm{ref}}_{\mathrm{ee}}(t) - \mathbf{p}_{\mathrm{ee}}(t)
        \right\|_2
        }{
        L_{\mathrm{arm}}
        }.
    \end{equation*}

    \item Maximum base tilt: This metric quantifies the largest roll–pitch attitude excursion during task execution.
    
    Let $\eta_{\mathrm{roll}}(t)$ and $\eta_{\mathrm{pitch}}(t)$ denote the base roll and pitch angles at timestep $t$. We define the maximum roll-pitch tilt magnitude over the rollout:
    \begin{equation*}
        U_{\mathrm{tilt}}
        =
        \max_{t \in \{1,\dots,N_t\}}
        \sqrt{
        \eta_{\mathrm{roll}}(t)^2 + \eta_{\mathrm{pitch}}(t)^2
        }.
    \end{equation*}
\end{itemize}
\subsection{Indoor and Outdoor Results}

\subsubsection{Indoor Experiments}

\begin{figure*}[t]
    \centering

    \begin{subfigure}[b]{0.32\textwidth}
        \centering
        \includegraphics[width=\linewidth]{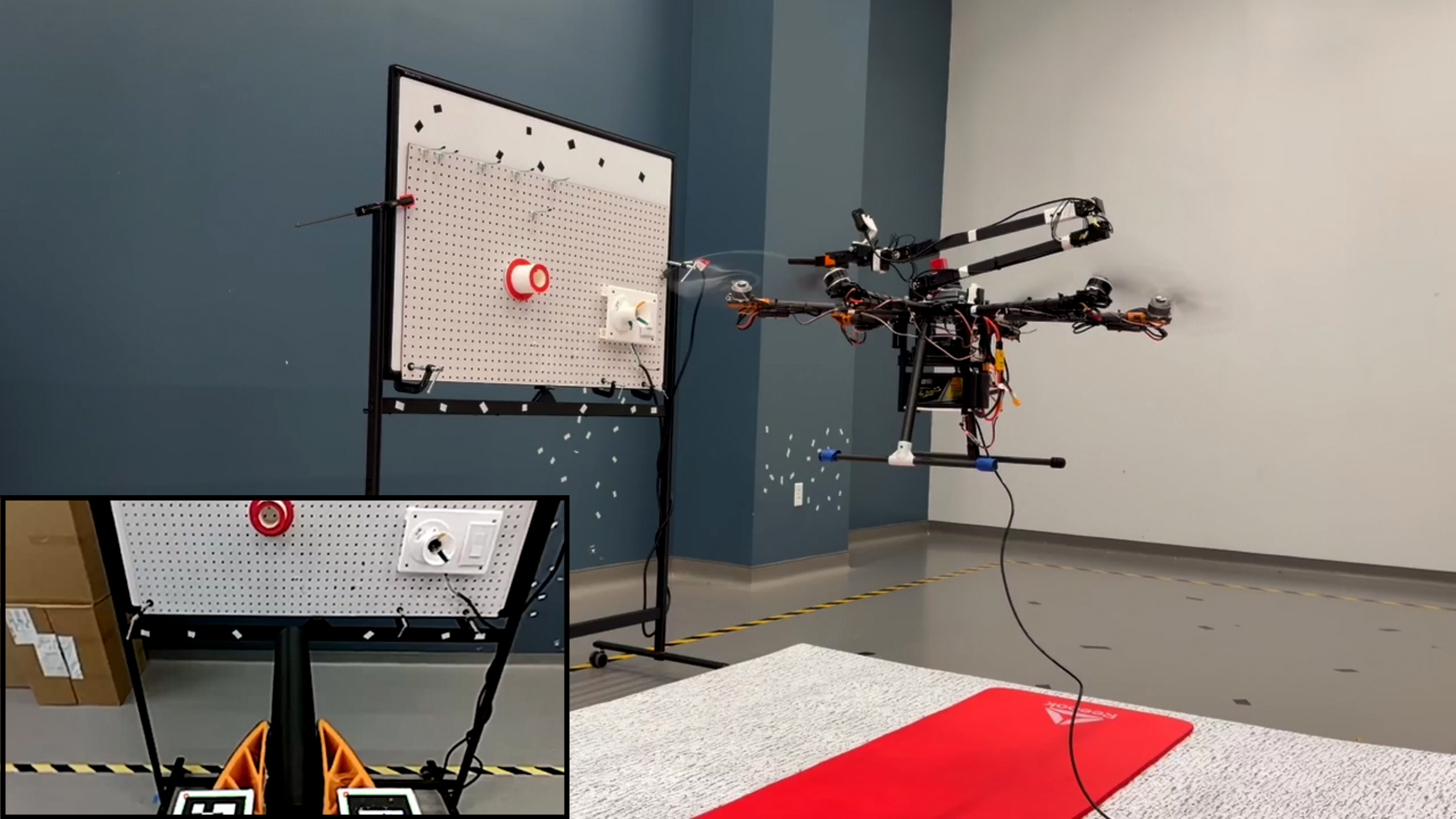}
        \caption{Policy inference initialization}
        \label{fig:indoor_init}
    \end{subfigure}
    \hfill
    \begin{subfigure}[b]{0.32\textwidth}
        \centering
        \includegraphics[width=\linewidth]{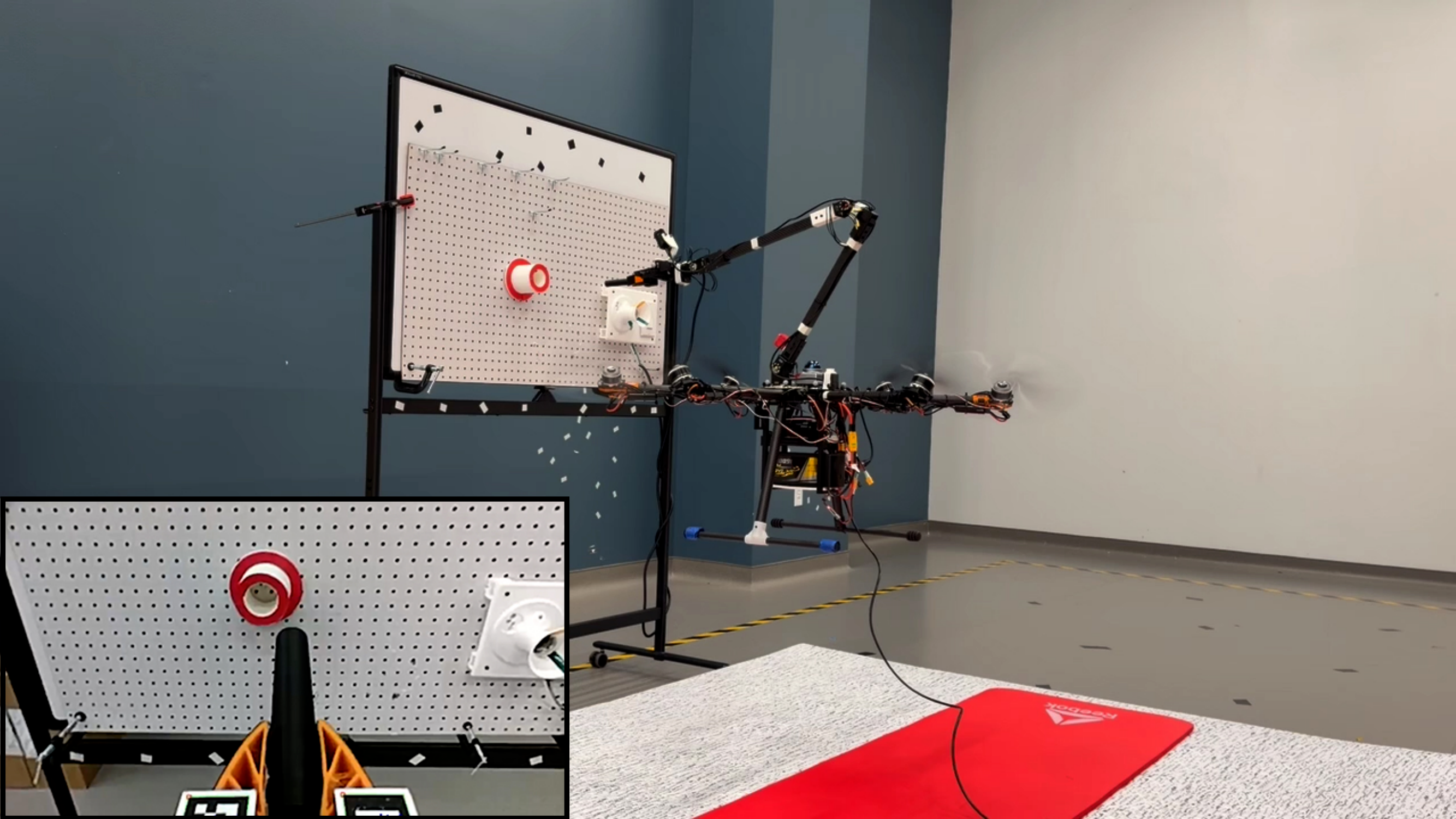}
        \caption{Approach to the target}
        \label{fig:indoor_approach}
    \end{subfigure}
    \hfill
    \begin{subfigure}[b]{0.32\textwidth}
        \centering
        \includegraphics[width=\linewidth]{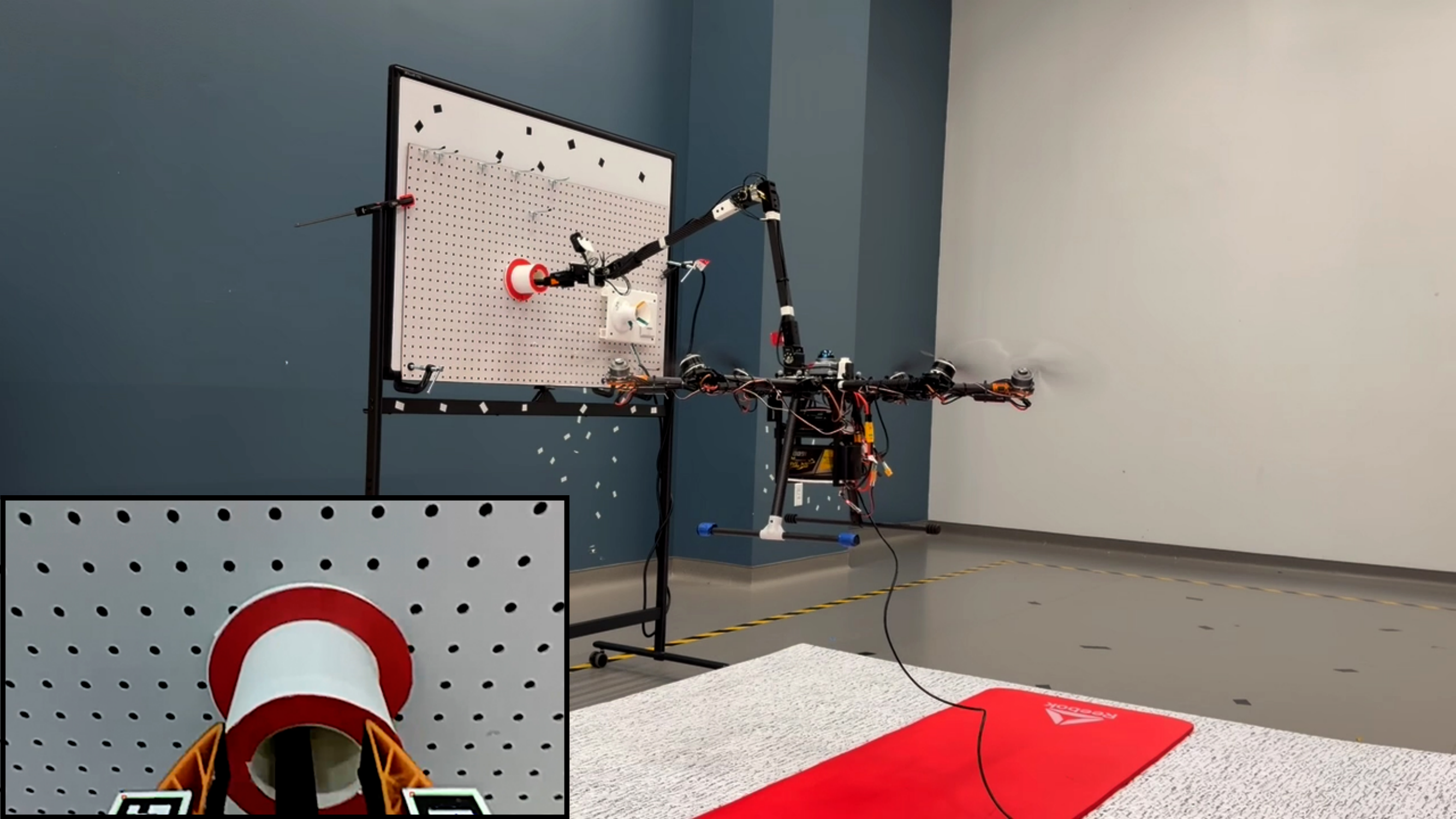}
        \caption{Peg insertion}
        \label{fig:indoor_insertion}
    \end{subfigure}

    \caption{Indoor peg-in-hole execution in the training environment, showing the initial configuration, target approach, and successful insertion. Insets show the onboard first person view used by the visuomotor policy.}
    \label{fig:indoor_peginhole}
\end{figure*}

\begin{figure*}[t]
    \centering

    \begin{subfigure}[b]{0.32\textwidth}
        \centering
        \includegraphics[width=\linewidth]{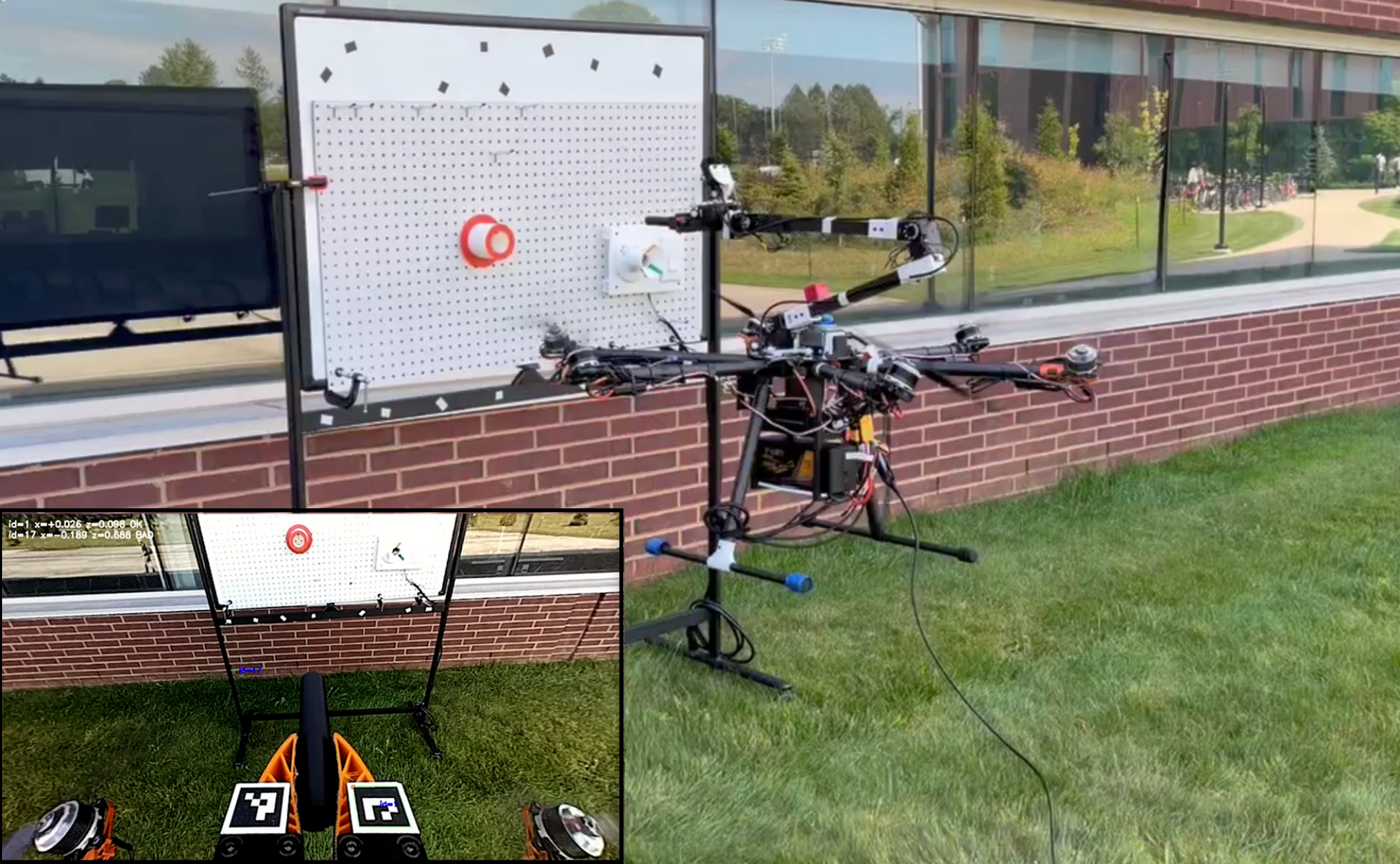}
        \caption{Policy inference initialization}
        \label{fig:outdoor_init}
    \end{subfigure}
    \hfill
    \begin{subfigure}[b]{0.32\textwidth}
        \centering
        \includegraphics[width=\linewidth]{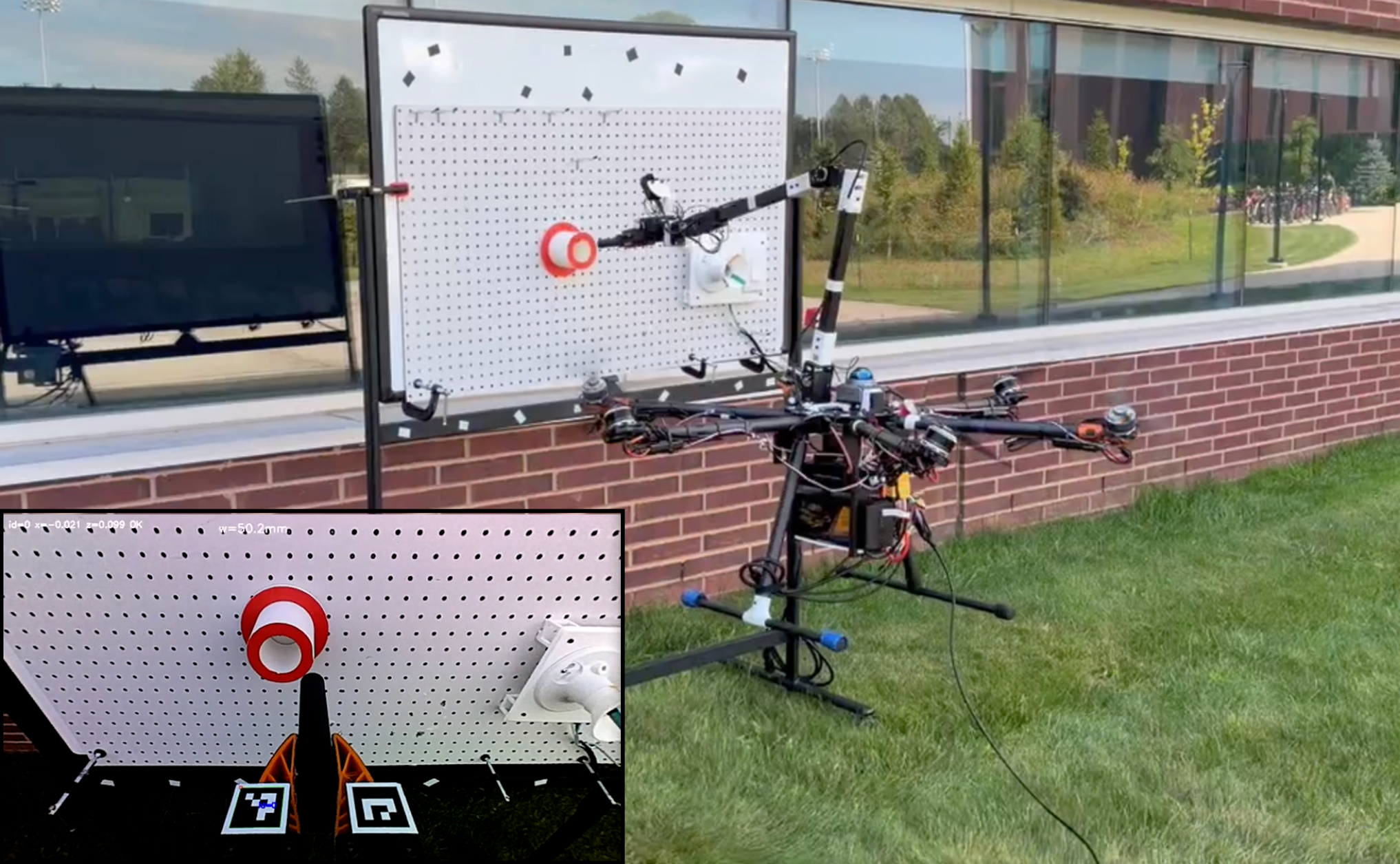}
        \caption{Approach to the target}
        \label{fig:outdoor_approach}
    \end{subfigure}
    \hfill
    \begin{subfigure}[b]{0.32\textwidth}
        \centering
        \includegraphics[width=\linewidth]{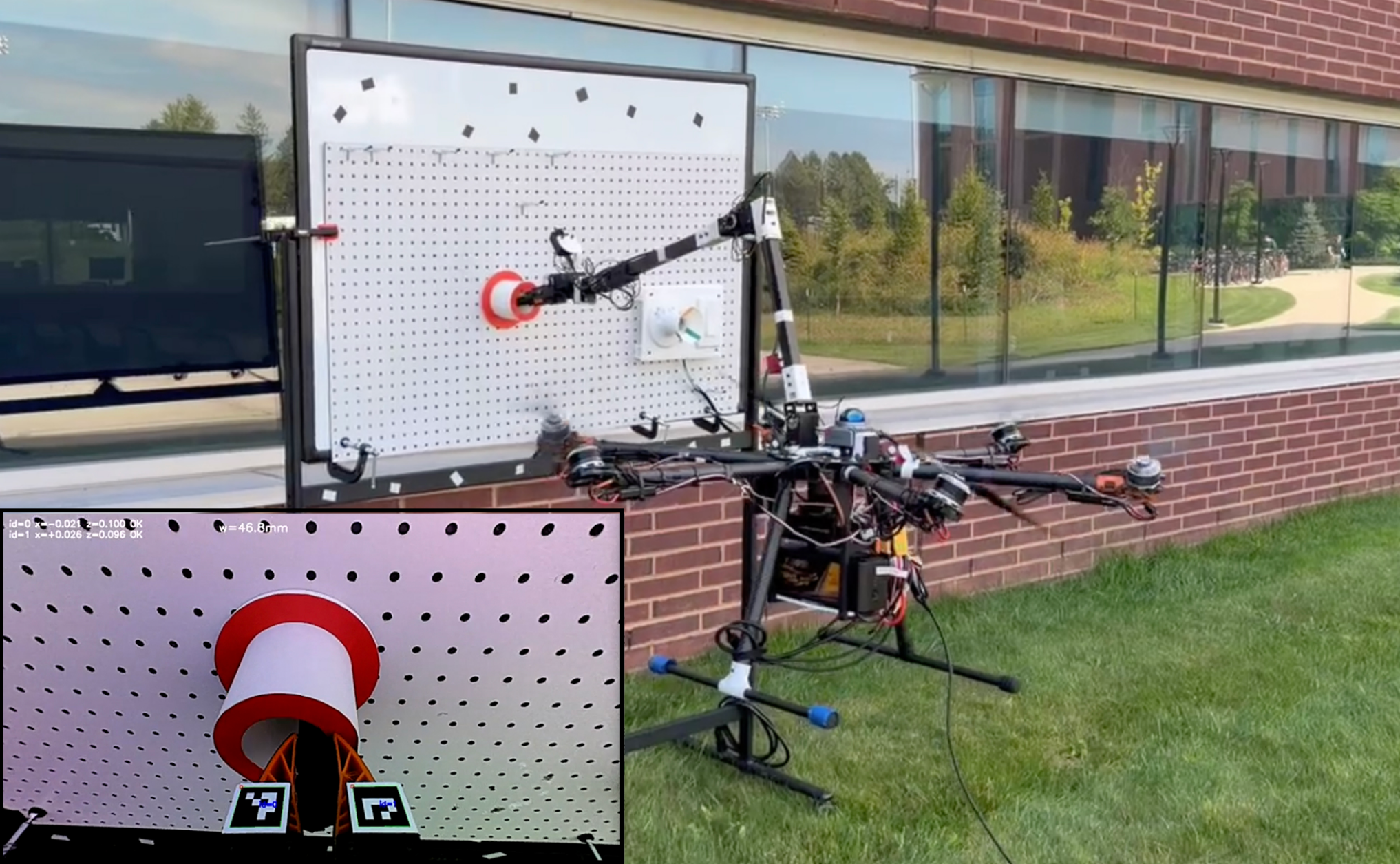}
        \caption{Peg insertion}
        \label{fig:outdoor_insertion}
    \end{subfigure}

    \caption{Outdoor peg-in-hole execution in an unseen environment using the same learned diffusion policy, with onboard LiDAR-inertial odometry for state estimation. Insets show the onboard first person view used by the visuomotor policy.}
    \label{fig:outdoor_peginhole}
    \vspace{-5mm}
\end{figure*}

We first evaluate the learned policy in the indoor environment. The system successfully completes all five peg-in-hole trials despite variations in the initial gripper configuration and target hole location. A representative trial is shown in Fig.~\ref{fig:indoor_peginhole}. The whole-body MPC achieves a normalized end-effector tracking error of $0.102$ while maintaining small vehicle tilt. 

\begin{table}[t]
    \centering
    \caption{End-effector tracking and base attitude performance in indoor and outdoor experiments.}
    \label{tab:controller_comparison}
    \begin{tabular}{lcc}
        \hline
        \textbf{Environment} & \textbf{Norm. EE Error} & \textbf{Max. Base Tilt [deg]}\\
        \hline
        Indoor & 0.102 & 3.32 \\
        Outdoor & 0.104 & 1.84 \\
        \hline
    \end{tabular}
    \vspace{-3mm}
\end{table} 

\begin{figure}[!t]
    \centering

    \begin{subfigure}{\linewidth}
        \centering
        \includegraphics[width=0.9\linewidth]{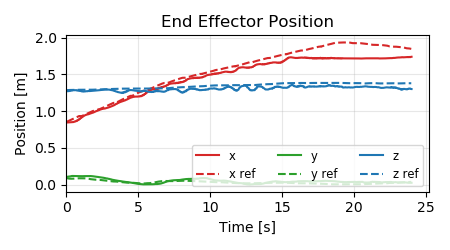}
        \label{fig:ee_tracking_indoor}
        \vspace{-3mm}
        \caption{Indoor}
    \end{subfigure}

    \begin{subfigure}{\linewidth}
        \centering
        \includegraphics[width=0.9\linewidth]{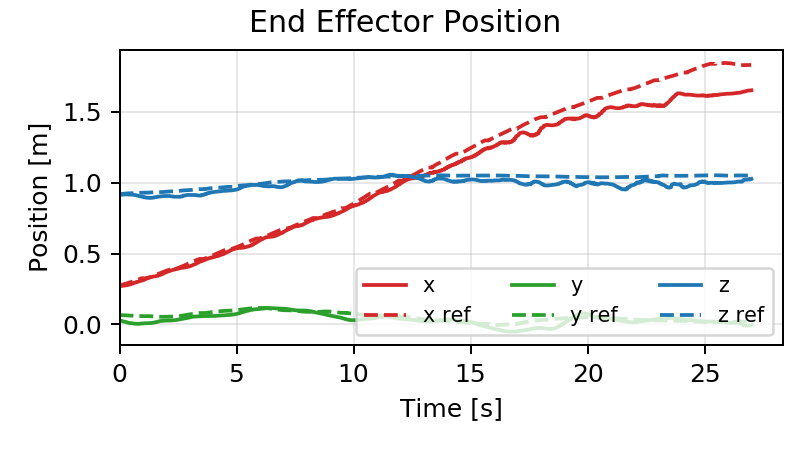}
        \label{fig:ee_tracking_outdoor}
        \vspace{-3mm}
        \caption{Outdoor}
    \end{subfigure}
    \caption{End-effector position tracking performance in both inddor and outdoor environments. Solid lines denote the measured end-effector position, while dashed lines denote the policy-generated reference trajectory.}
    \label{fig:ee_tracking}
    \vspace{-5mm}
\end{figure}

Fig.~\ref{fig:ee_tracking} shows representative end-effector tracking performance under whole-body MPC. The controller coordinates the aerial base and manipulator to track the policy-generated trajectory while maintaining stable vehicle motion. These indoor experiments establish the baseline performance of the learned policy before outdoor evaluation.

% \begin{figure}[!t]
%     \centering

%     \begin{subfigure}{\linewidth}
%         \centering
%         \includegraphics[width=0.9\linewidth]{Figures/IK_EE.png}
%         % \caption{End-effector tracking.}
%         \label{fig:ee_tracking_indoor}
%     \end{subfigure}

%     \vspace{-3mm}

%     \begin{subfigure}{\linewidth}
%         \centering
%         \includegraphics[width=0.9\linewidth]{Figures/MPC_EE.png}
%         % \caption{Aerial-base motion.}
%         \label{fig:base_tracking_indoor}
%     \end{subfigure}
%     \vspace{-7mm}

%     \caption{End-effector position tracking during an indoor Diffusion Policy rollout using the IK-based controller (top) and whole-body MPC (bottom). Solid lines denote the measured end-effector position, while dashed lines denote the policy-generated reference trajectory.}
%     \label{fig:indoor_tracking}
%     \vspace{-8mm}
% \end{figure}

\subsubsection{Outdoor Experiments}

We next evaluate the same learned policy in an unseen outdoor environment to assess its ability to generalize beyond the indoor demonstration setting. During outdoor deployment, the aerial platform relies entirely on onboard LIO for floating-base state estimation, while the learned Diffusion Policy and low-level controller remain unchanged.

We conduct three outdoor peg-in-hole trials, and the system successfully completes all three trials without human intervention after policy execution begins. A representative trial is shown in Fig.~\ref{fig:outdoor_peginhole}. Starting from an initial configuration different from those observed during data collection, the policy guides the aerial manipulator toward the target and generates the final insertion motion required to place the stick into the hole.

Table~\ref{tab:controller_comparison} shows comparable normalized end-effector tracking errors for indoor and outdoor experiments. Despite replacing motion-capture system with onboard LIO during outdoor deployment, the whole-body controller maintains similar trajectory-tracking performance. The maximum base tilt also remains below $4^\circ$ in both environments, indicating that the manipulation trajectory is executed without excessive vehicle attitude excursion.

The outdoor experiments show that a visuomotor policy trained solely from indoor demonstrations can generalize to a visually and geometrically different environment without retraining, while onboard LiDAR–inertial odometry provides sufficiently stable state feedback for execution outside a motion-capture environment. Together, these results demonstrate the feasibility of transferring demonstration-driven aerial manipulation from controlled indoor settings to outdoor deployment.
\section{Conclusion}
\label{sec: conclusion}

This work presented an outdoor aerial manipulation framework integrating Diffusion Policy, onboard LiDAR--inertial state estimation, and whole-body MPC. Experiments on a physical aerial manipulator demonstrated that a policy trained entirely from indoor demonstrations can be deployed in an unseen outdoor environment without external motion capture, outdoor demonstrations, or policy retraining. The results show that combining transferable visuomotor policies with onboard localization and coordinated whole-body control provides a practical pathway toward aerial manipulation beyond controlled laboratory environments. Future work will evaluate the framework on more diverse manipulation tasks and challenging outdoor conditions.
\section*{ACKNOWLEDGMENT}

This work was supported in part by the National Science Foundation under Awards No. 2541976. Any opinions, findings, and conclusions or recommendations expressed in this paper are those of the author and do not necessarily reflect the views of the National Science Foundation.

\bibliographystyle{IEEEtran}
\bibliography{reference}

@ARTICLE{ruggiero2018aerial,
  author={Ruggiero, Fabio and Lippiello, Vincenzo and Ollero, Anibal},
  journal={IEEE Robotics and Automation Letters}, 
  title={Aerial Manipulation: A Literature Review}, 
  year={2018},
  volume={3},
  number={3},
  pages={1957-1964},
  doi={10.1109/LRA.2018.2808541}}

@ARTICLE{ollero2021past,
  author={Ollero, Anibal and Tognon, Marco and Suarez, Alejandro and Lee, Dongjun and Franchi, Antonio},
  journal={IEEE Transactions on Robotics}, 
  title={Past, Present, and Future of Aerial Robotic Manipulators}, 
  year={2022},
  volume={38},
  number={1},
  pages={626-645},
  doi={10.1109/TRO.2021.3084395}}

@article{xu2022fastlio2,
  author  = {Xu, Wei and Cai, Yixi and He, Dongjiao and Lin, Jiarong and Zhang, Fu},
  title   = {FAST-LIO2: Fast Direct LiDAR-Inertial Odometry},
  journal = {IEEE Transactions on Robotics},
  year    = {2022},
  volume  = {38},
  number  = {4},
  pages   = {2053--2073}
}

@inproceedings{chi2023diffusion,
  author    = {Chi, Cheng and Feng, Siyuan and Du, Yilun and Xu, Zhenjia and Cousineau, Eric and Burchfiel, Benjamin C. M. and Song, Shuran},
  title     = {Diffusion Policy: Visuomotor Policy Learning via Action Diffusion},
  booktitle = {Proceedings of Robotics: Science and Systems},
  year      = {2023},
  doi       = {10.15607/RSS.2023.XIX.026}
}

@inproceedings{he2025flying,
  author    = {He, Guanqi and Guo, Xiaofeng and Tang, Luyi and Zhang, Yuanhang and Mousaei, Mohammadreza and Xu, Jiahe and Geng, Junyi and Scherer, Sebastian and Shi, Guanya},
  title     = {Flying Hand: End-Effector-Centric Framework for Versatile Aerial Manipulation Teleoperation and Policy Learning},
  booktitle = {Proceedings of Robotics: Science and Systems},
  year      = {2025},
  doi       = {10.15607/RSS.2025.XXI.130},
  eprint    = {2504.10334},
  archivePrefix = {arXiv}
}

@inproceedings{zhao2023act,
  author    = {Zhao, Tony Z. and Kumar, Vikash and Levine, Sergey and Finn, Chelsea},
  title     = {Learning Fine-Grained Bimanual Manipulation with Low-Cost Hardware},
  booktitle = {Proceedings of Robotics: Science and Systems},
  year      = {2023},
  doi       = {10.15607/RSS.2023.XIX.016}
}

@inproceedings{ha2024umilegs,
  author    = {Ha, Huy and Gao, Yihuai and Fu, Zipeng and Tan, Jie and Song, Shuran},
  title     = {{UMI}-on-Legs: Making Manipulation Policies Mobile with Manipulation-Centric Whole-Body Controllers},
  booktitle = {Proceedings of the 8th Conference on Robot Learning},
  year      = {2025},
  volume    = {270},
  series    = {Proceedings of Machine Learning Research},
  pages     = {5254--5270},
  publisher = {PMLR}
}

@article{gupta2025umionair,
  author  = {Gupta, Harsh and Guo, Xiaofeng and Ha, Huy and Pan, Chuer and Cao, Muqing and Lee, Dongjae and Scherer, Sebastian and Song, Shuran and Shi, Guanya},
  title   = {{UMI}-on-Air: Embodiment-Aware Guidance for Embodiment-Agnostic Visuomotor Policies},
  journal = {arXiv preprint arXiv:2510.02614},
  year    = {2025},
  eprint  = {2510.02614},
  archivePrefix = {arXiv}
}

@article{levine2016endtoend,
  author  = {Levine, Sergey and Finn, Chelsea and Darrell, Trevor and Abbeel, Pieter},
  title   = {End-to-End Training of Deep Visuomotor Policies},
  journal = {Journal of Machine Learning Research},
  year    = {2016},
  volume  = {17},
  number  = {39},
  pages   = {1--40}
}

@inproceedings{yang2014dynamics,
  author    = {Yang, Hyunsoo and Lee, Dongjun},
  title     = {Dynamics and Control of Quadrotor with Robotic Manipulator},
  booktitle = {2014 IEEE International Conference on Robotics and Automation},
  year      = {2014},
  pages     = {5544--5549},
  doi       = {10.1109/ICRA.2014.6907674}
}

@article{kim2018cooperative,
  author  = {Kim, Suseong and Seo, Hoseong and Shin, Jongho and Kim, H. Jin},
  title   = {Cooperative Aerial Manipulation Using Multirotors With Multi-{DOF} Robotic Arms},
  journal = {IEEE/ASME Transactions on Mechatronics},
  year    = {2018},
  volume  = {23},
  number  = {2},
  pages   = {702--713},
  doi     = {10.1109/TMECH.2018.2792318}
}

@article{ramonsoria2020outdoor,
  author  = {Ram{\'o}n-Soria, Pablo and Arrue, Bego{\~n}a C. and Ollero, An{\'i}bal},
  title   = {Grasp Planning and Visual Servoing for an Outdoors Aerial Dual Manipulator},
  journal = {Engineering},
  year    = {2020},
  volume  = {6},
  number  = {1},
  pages   = {77--88},
  doi     = {10.1016/j.eng.2019.11.003}
}

@article{ubellacker2024highspeed,
  author  = {Ubellacker, Samuel and Ray, Aaron and Bern, James M. and Strader, Jared and Carlone, Luca},
  title   = {High-Speed Aerial Grasping Using a Soft Drone with Onboard Perception},
  journal = {npj Robotics},
  year    = {2024},
  volume  = {2},
  pages   = {5},
  doi     = {10.1038/s44182-024-00012-1}
}

@inproceedings{bauer2025wild,
  author    = {Bauer, Erik and Bl{\"o}chlinger, Marc and Strauch, Pascal and Raayatsanati, Arman and Curdin, Cavelti and Katzschmann, Robert K.},
  title     = {An Open-Source Soft Robotic Platform for Autonomous Aerial Manipulation in the Wild},
  booktitle = {Proceedings of the 8th Conference on Robot Learning},
  year      = {2025},
  volume    = {270},
  series    = {Proceedings of Machine Learning Research},
  pages     = {3094--3106}
}

@inproceedings{zhan2026contactaware,
  author        = {Zhan, Yuanzhu and Jiang, Yufei and Cao, Muqing and Geng, Junyi},
  title         = {Aerial Manipulation with Contact-Aware Onboard Perception and Hybrid Control},
  booktitle     = {2026 IEEE International Conference on Robotics and Automation (ICRA)},
  year          = {2026},
  eprint        = {2602.08251},
  archivePrefix = {arXiv},
  primaryClass  = {cs.RO}
}

@inproceedings{jiang2025selfsupervised,
  author    = {Jiang, Yufei and Zhan, Yuanzhu and Gupta, Harsh Vardhan and Borde, Chinmay and Geng, Junyi},
  title     = {A Self-Supervised Learning Approach with Differentiable Optimization for UAV Trajectory Planning},
  booktitle = {2026 IEEE International Conference on Robotics and Automation (ICRA)},
  year      = {2026}
}

@INPROCEEDINGS{chi2024umi, 
    AUTHOR    = {Cheng Chi AND Zhenjia Xu AND Chuer Pan AND Eric Cousineau AND Benjamin Burchfiel AND Siyuan Feng AND Russ Tedrake AND Shuran Song}, 
    TITLE     = {{Universal Manipulation Interface: In-The-Wild Robot Teaching Without In-The-Wild Robots}}, 
    BOOKTITLE = {Proceedings of Robotics: Science and Systems}, 
    YEAR      = {2024}, 
    ADDRESS   = {Delft, Netherlands}, 
    MONTH     = {July}, 
    DOI       = {10.15607/RSS.2024.XX.045} 
}

@InProceedings{ronneberger2015unet,
author="Ronneberger, Olaf
and Fischer, Philipp
and Brox, Thomas",
editor="Navab, Nassir
and Hornegger, Joachim
and Wells, William M.
and Frangi, Alejandro F.",
title="U-Net: Convolutional Networks for Biomedical Image Segmentation",
booktitle="Medical Image Computing and Computer-Assisted Intervention -- MICCAI 2015",
year="2015",
publisher="Springer International Publishing",
address="Cham",
pages="234--241",
}

@inproceedings{
song2021DDIM,
title={Denoising Diffusion Implicit Models},
author={Jiaming Song and Chenlin Meng and Stefano Ermon},
booktitle={International Conference on Learning Representations},
year={2021},
url={https://openreview.net/forum?id=St1giarCHLP}
}

\end{document}